\documentclass[10pt, a4paper]{article}

\usepackage{lrec2026}
\usepackage{spverbatim}
\usepackage{amsmath}
\usepackage[title]{appendix}
\usepackage{graphicx}
\usepackage[figuresleft]{rotating}
\usepackage{subcaption}
\usepackage{multirow}
\usepackage{afterpage}

\title{MultiZebraLogic: A Multilingual Logical Reasoning Benchmark}

\name{Sofie Helene Bruun, Dan Saattrup Smart}

\address{
    The Alexandra Institute \\
    Rued Langgaards Vej 7, 5D, 2300 Copenhagen S, Denmark \\
    \{sofie.bruun, dan.smart\}@alexandra.dk\\
}

\abstract{
    We create high-quality datasets for LLM evaluation of logical reasoning skills
    across nine different languages, which have been manually checked by fluent
    speakers. The datasets consist of so-called zebra puzzles, and we analyse different
    ways of tuning the difficulty of the puzzles to fit modern LLMs. This includes the
    size of the puzzle (number of objects and number of clues), as well as a novel
    addition of red herring clues containing only irrelevant information. We show that
    presence of red herrings indeed makes the puzzles significantly harder for the
    models, and we find puzzle sizes 2×3 and 4×5 are sufficiently challenging for GPT-4o
    mini (a non-reasoning model) and o3-mini (a reasoning model), respectively. We
    analyse whether LLM performance of these are sensitive to the language, the cultural
    sensitivity of the puzzle theme, and the choice of clue types. These analyses are
    conducted with English and Danish, where we show that there is no significant
    difference for either of these three aspects, at least for the OpenAI models GPT-4o
    mini and o3-mini, chosen as representative non-reasoning and reasoning models,
    respectively. We publish the datasets for each of the nine languages for the
    identified sizes 2×3 and 4×5. We also publish the code used to generate the puzzles,
    which can be used to extend the benchmark into more languages.
  \\
  \newline
}

\begin{document}

\maketitleabstract

\section{Introduction}
\label{sec:introduction}

With the advent of large language models (LLMs) with reasoning capabilities, evaluating
their logical reasoning skills is essential. Existing reasoning datasets focus solely on
\textit{common-sense reasoning}
\citelanguageresource{zellers2019hellaswag,lin2021common,ponti2020xcopa} or English-only
tasks
\citelanguageresource{lin2025zebralogic,chenenigmata,patel2024multi,wei2025satbench}.

To remedy this, we create \textit{MultiZebraLogic}, a multilingual logical reasoning
benchmark using zebra puzzles \citep{zebra_history}. First published in 1962, these 
puzzles require multi-step reasoning: the solver is given objects with attributes and 
clues describing relationships between them\footnote{The original question was ``Who 
owns the zebra?'', which has named the puzzle genre.}, finding a solution that 
satisfies all clues (see Figure~\ref{fig:simple_puzzle}).

This type of constraint satisfaction problem is easy to generate and requires multiple
steps to solve. Simple algorithms for solving zebra puzzles exist, but to follow them,
most humans would need to draw diagrams of excluded combinations. Assuming that such
diagrams are allowed, humans can thus solve any zebra puzzle, albeit slowly.

In building the benchmark we identify appropriate sizes (number of objects and number of
attributes per object) for LLM evaluation, and also examine other ways of increasing
difficulty by adding red herrings (non-informative clues), more clue types, and a
culture-specific theme: Danish smørrebrød (open sandwiches) with different ingredients.

Our main contributions are:

\begin{itemize}
    \item A multilingual logical reasoning
      benchmark\footnote{\url{https://huggingface.co/datasets/alexandrainst/zebra\_puzzles}}
      designed for both reasoning and non-reasoning LLMs, covering 9
      Germanic languages\footnote{English, Danish, Swedish, Norwegian Bokmål, Norwegian
      Nynorsk, Faroese, Icelandic, German and Dutch.}.
    \item Source code for puzzle generation built for scalability to more languages or
      themes \footnote{\url{https://github.com/alexandrainst/zebra\_puzzles}}.
    \item Analysis of effects on puzzle difficulty from red herrings (non-informative
      clues), a culture-specific theme, clue types, and a medium vs. high resource
      language.
\end{itemize}

\begin{figure}
\scriptsize
\begin{verbatim}
A row of houses have numbers 1 to 2 from left to
right.

In each house lives a person with unique attributes
in each of the following categories:

Jobs: nurse and police officer.
Favourite book genres: fantasy and romance.
Hobbies: bouldering and handball.

We also know the following:

1. The person who plays handball knows that snails
   are molluscs.
2. The police officer lives to the left of the nurse.
3. The person who plays handball does not live in
   house no. 2.
4. The romance reader lives in house no. 2.
5. The person with glasses does not live in house
   no. 1.

Who has which attributes and lives in which house?

Please submit your answer as a JSON dictionary in
the format below. Each row must begin with object_X
where X is the house number. Each column represents
a category, and they should be in the same order
as in the list of categories above.

{
    "object_1": [
        "jobs_1",
        "favourite book genres_1",
        "hobbies_1"
    ],
    "object_2": [
        "jobs_2",
        "favourite book genres_2",
        "hobbies_2"
    ]
}
\end{verbatim}
\caption{
  A zebra puzzle with 2 objects and 3 attributes for each object (2x3). Two red
  herrings are also included in the list of clues. See Table~\ref{tab:solution_example}
  for the solution.
}
\label{fig:simple_puzzle}
\end{figure}

\section{Related Work}
\label{sec:related_work}

A wide range of benchmarks has been developed to systematically evaluate LLMs’ logical
reasoning skills across different reasoning types and complexities.

\citetlanguageresource{lin2025zebralogic} built a ``ZebraLogic'' benchmark to measure
how logical reasoning performance of LLMs scale with zebra puzzle complexity in English.
Our puzzle generation approach will be similar, but we attempt to increase difficulty
for all puzzle sizes by adding more clue types, more languages as well as red herrings
(non-informative clues).

Other benchmarks focus on specific reasoning domains. LogicBench
\citelanguageresource{parmar2024logicbench} targets 25 inference rules from
propositional to non-monotonic logics, while JustLogic
\citelanguageresource{chen2025justlogic} emphasises deductive reasoning. SATBench
\citelanguageresource{wei2025satbench} challenges LLMs with search-based logical puzzles
from Boolean satisfiability problems, and KOR-Bench \citelanguageresource{ma2025kor}
evaluates knowledge-orthogonal reasoning.

\section{Methodology}
\label{sec:methodology}

\subsection{Puzzle Generation}
\label{sec:puzzle_generation}

For a given theme and language, we generate puzzles with the following structure:

\begin{enumerate}
    \item Introduction to the theme and rules including the number of objects,
      $N_{\mathrm{objects}}$, and attributes per object, $N_{\mathrm{attributes}}$.
    \item A list of possible attributes and their categories.
    \item A list of clues and red herrings.
    \item Instructions on how to format the solution.
\end{enumerate}

Objects could be houses, and attributes belong to categories such as jobs and pets.
Multiple phrases\footnote{E.g. ``the baker'', ``is a baker'' and ``is not a baker''.}
are included per attribute to fit different sentence structures without adding
language-specific grammatical rules.

We start by generating solutions by randomly sampling categories and attributes within
each category for each object, from a fixed list of categories and attributes.  We also
assign each row an object index. See Table~\ref{tab:solution_example} for an example of
a solution.

\begin{table}[]
    \centering
    \begin{tabular}{c|c|c|c}
         object\_1 & police officer & fantasy & handball \\
         \hline
         object\_2 & nurse & romance & bouldering
    \end{tabular}
    \caption{
      Example of a $N_{\mathrm{objects}} \times (N_{\mathrm{attributes}}+1)$ solution
      matrix for a 2×3 puzzle in the English houses theme. Each object represents a
      house and its row lists the attributes of the resident. See
      Figure~\ref{fig:simple_puzzle} for the corresponding puzzle.
    }
    \label{tab:solution_example}
\end{table}

To generate a clue, we sample a clue type from Table~\ref{tab:clue_types} and sample
solution objects from the previous step along with attributes meeting the constraints of
the clue. If the presented attribute order is irrelevant, attributes are sorted by
category in the order that would typically sound the most natural\footnote{E.g., ``The
nurse loves oranges.'' instead of ``The person who loves oranges is a nurse.''}.
Appendix~\ref{app:clue_examples} shows full clue examples.

Using the Python \texttt{constraint} package \citep{python-constraint}, we define a
constraint satisfaction problem per puzzle and solve it. If a suggested clue changes the
number of possible solutions, we keep it and iterate until a unique solution remains.
Then, we remove each clue and only re-add it if the solution degenerates. This causes a
bias towards including more informative clues, as illustrated in
Appendix~\ref{app:clue_frequency}.

\afterpage{
\begin{table*}[]
    \centering
    \footnotesize
    \begin{tabular}{l|c|c}
        Clue type & Positional constraint & Requirement \\
        \hline
        \texttt{found\_at} & $X = P$ &  \\
        \texttt{not\_at} & $X \neq P$ &  \\
        \texttt{same\_object} & $X = Y$ & $N_{\mathrm{attributes}}>1$\\
        \texttt{not\_same\_object} & $X \neq Y$ & $N_{\mathrm{attributes}}>1$\\
        \texttt{next\_to} & $|X-Y|=1$ & $N_{\mathrm{objects}}>2$ \\
        \texttt{not\_next\_to} & $|X-Y|>1$ & $N_{\mathrm{objects}}>2$ \\
        \texttt{just\_left\_of} & $Y-X=1$ & $N_{\mathrm{objects}}>2$ \\
        \texttt{just\_right\_of} & $X-Y=1$ & $N_{\mathrm{objects}}>2$ \\
        \texttt{left\_of} & $X<Y$ &  \\
        \texttt{right\_of} & $X>Y$ &  \\
        \texttt{between} & $X<Y<Z \lor X>Y>Z$ & $N_{\mathrm{objects}}>2$ \\
        \texttt{not\_between} &
          $\neg(X<Y<Z \lor X>Y>Z)\land X\neq Y\land X\neq Z\land Y\neq Z$ &
          $N_{\mathrm{objects}}>2$ \\
        \texttt{one\_between} & $|X-Y|=2$ & $N_{\mathrm{objects}}>2$ \\
        \texttt{multiple\_between} &
          $|X-Y|=N_{\mathrm{between}}+1$ & $N_{\mathrm{objects}}>3$

    \end{tabular}
    \caption{
      List of clue types and their positional constraints of objects $X$, $Y$ and $Z$.
      $P$ is a specific position, and $N_{\mathrm{between}}$ is the number of objects
      between $A$ and $B$. Requirements are mentioned when they are stricter than the
      general puzzle generation requirements ($N_{\mathrm{objects}}>1,
      N_{\mathrm{attributes}}>0$). When multiple clue types would reveal the same
      information, the requirements exclude one for improved
      naturalness\protect\footnotemark.
    }
    \label{tab:clue_types}
\end{table*}
}

Each red herring mentions either one of the attributes present in the solution, or none
at all. We include 8 types; some follow the same templates as real clues, while others
are new, such as random facts. We shuffle the order of clues and red herrings. See
Figure~\ref{fig:simple_puzzle} and Appendix~\ref{app:advanced_puzzle} for examples of
puzzles and all clue and red herring types.

Red herrings require less effort, as they contain no useful information. Some red
herring types follow the same templates as real clues but with irrelevant attributes.
Other types are new such as randomly chosen facts or statements about friendship related
to objects. We end by shuffling the order of clues and red herrings.

\subsubsection{Translation}
\label{sec:translation}

The priorities for linguistic puzzle components are: 1) Correctness. Text must be
linguistically acceptable. 2) Unambiguity. Clues must represent a unique solution. 3)
Naturalness. Phrases should sound typical of the chosen language. 4) Ease of generation.
Puzzle generation should be simple. 5) Consistency. Text should be consistent in meaning
and form across languages. 6) Diversity. A variety of properties and clue types should
be included. There are tradeoffs between priorities\footnote{For unambiguity, we prefer
``There are $n$ houses between $X$ and $Y$'' although ``$X$ lives $n$ houses away from
$Y$'' is slightly more natural. In Icelandic, for ``$X$ does not like $H$'' we use ``$X$
elskar ekki $H$'' instead of ``$X$ líkar ekki $H$'' to avoid the dative case for $X$ --
this simplifies generation at a small cost to naturalness and consistency.}.

Translation to new Germanic languages requires few changes to the puzzle generation
algorithm itself, as we mostly avoid grammatical and social gender. The most important
difference lies in the use of grammatical cases for attributes and clue types in
Faroese, Icelandic and German. In German and Dutch, we add more forms of some clauses,
to place the verb at the end of subordinate clauses. Some phrases are directly replaced
after initial puzzle generation, such as the combination of ``von dem'' into ``vom'' in
German.

All translations are drafted by the authors and reviewed by native/fluent speakers. For
the drafts, we use Google Translate \citep{google_translate}, dictionaries
\citep{svenska, wordify, ordbøkene, divvun}, suggestions from GitHub Copilot with
GPT-4.1 \citep{copilot,gpt-41} and Wikipedia \citep{wikipedia}.

\subsection{Evaluating LLM Performance}
\label{sec:evaluation}

We explore puzzle difficulty for two LLMs. To represent a reasoning model, we choose
o3-mini \citep{o3-mini} with \texttt{max\_completion\_tokens} set to 100,000 and
\texttt{reasoning\_effort} set to ``medium''. As a non-reasoning model, we select GPT-4o
mini \citep{gpt-4o} with \texttt{max\_completion\_tokens} set to 16,384 and
\texttt{temperature} set to 0, to ensure reproducible evaluations\footnote{We use a
larger \texttt{max\_completion\_tokens} value for o3-mini to account for the reasoning
trace.}. They should output a JSON response for each puzzle, which is compared to the
solution. See Appendix~\ref{app:evaluation} for more details.

We use datasets of 100 puzzles per size with the smørrebrød theme, and evaluate using
all sizes from 2×1 to 5×5, except 5×4 and 5×5 ($N_{\mathrm{objects}}\times
N_{\mathrm{attributes}}$), as larger puzzles would take too many resources for both
generation and evaluation. Puzzles with 1 object would require no clues. We generate 5
red herrings per puzzle and remove 4 or 5 to also create datasets with one or no red
herring.

Performance is evaluated using the metrics of \citetlanguageresource{lin2025zebralogic}:
Puzzle-level accuracy, $A_{\mathrm{puzzle}}$, which is 1 for a correct response and 0
otherwise; and cell-wise accuracy, $A_{\mathrm{cell}}$, which is the fraction of correct
cells in the response matrix.

We compute standard deviations assuming that $A_{\mathrm{puzzle}}$ follows a Bernoulli
distribution and $A_{\mathrm{cell}}$ approximately follows a normal distribution. See
Appendix~\ref{app:std} for more explanation of the use of standard deviations.

\section{Results}
\label{sec:results}

\begin{table*}[htb!]
    \centering
    \begin{tabular}{ll|c|c|c}
         & & Danish smørrebrød & Danish houses & English houses \\
        \hline
        \multirow{2}{*}{$A_{\mathrm{puzzle}}$} &
          Mean & 0.42$\pm$0.05 & 0.33$\pm$0.05 & 0.40$\pm$0.05 \\
        & Sample standard deviation & 0.5 & 0.5 & 0.5 \\
        \hline
        \multirow{2}{*}{$A_{\mathrm{cell}}$} &
          Mean & 0.66$\pm$0.04 & 0.66$\pm$0.04 & 0.67$\pm$0.04 \\
        & Sample standard deviation & 0.4 & 0.4 & 0.4
    \end{tabular}
    \caption{
      Comparison of o3-mini performance on 4×5 puzzles with 5 red herrings in the Danish
      smørrebrød, Danish houses and English houses themes (100 of each). Standard errors
      are included for mean values. Performance does not vary significantly by theme.
    }
    \label{tab:theme_scores}
\end{table*}

\subsection{Model Comparison}
\label{sec:model_comparison}

\begin{figure*}[h]
	\centering
    \resizebox{0.41\hsize}{!}{\includegraphics[trim={0.31cm 1.46cm 0 0.4cm},clip]{
      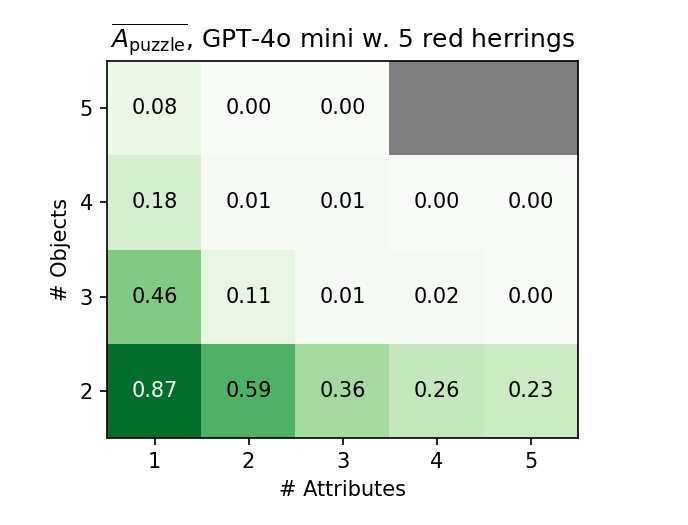}
    }
    \hspace{1.82cm}
    \resizebox{0.355\hsize}{!}{\includegraphics[trim={1.8cm 1.46cm 0 0.4cm},clip]{
      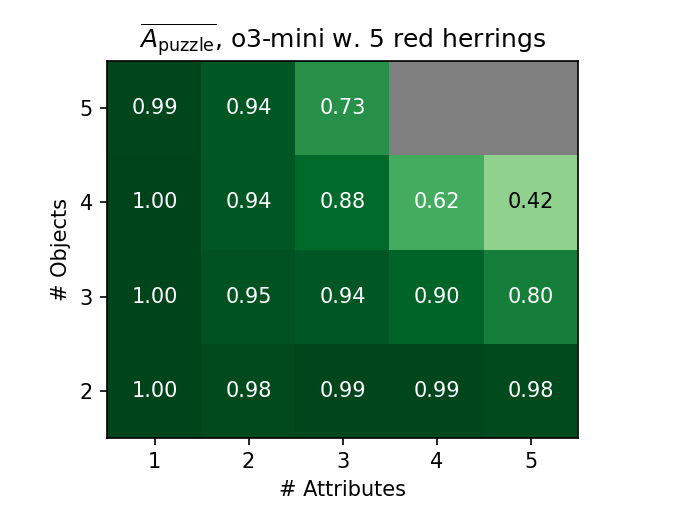}
    }
    \hspace{10cm}
    \resizebox{0.478\hsize}{!}{\includegraphics[trim={0 0.49cm 0 0},clip]{
      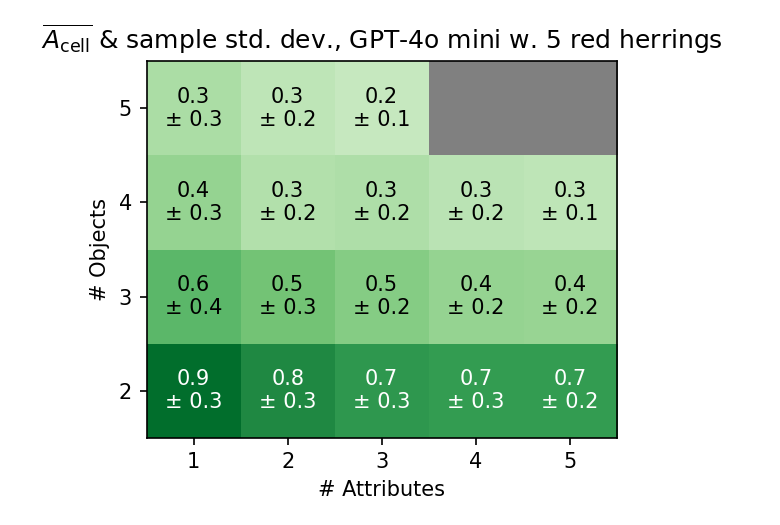}
    }
    \resizebox{0.478\hsize}{!}{\includegraphics[trim={0cm 0.49cm 0 0},clip]{
      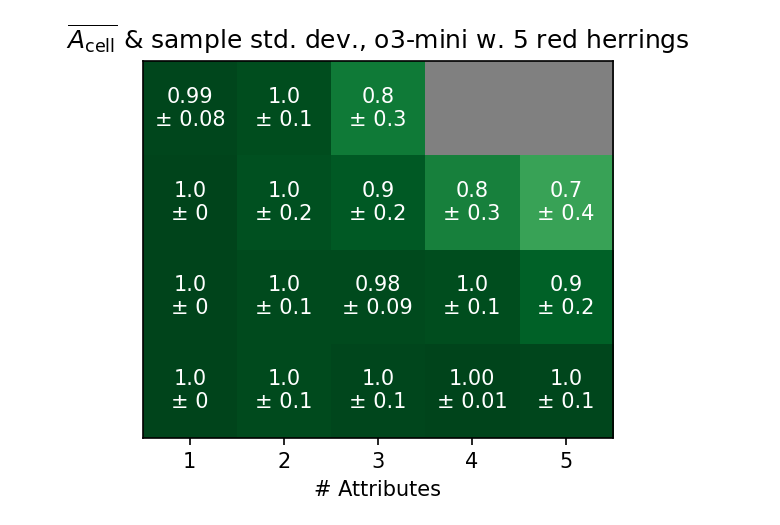}
    }
  \caption{
    $\overline{A_{\mathrm{puzzle}}}$ (upper row) and $\overline{A_{\mathrm{cell}}}$
    (lower row) for GPT-4o mini (left column) and o3-mini (right column) for 100 puzzles
    with 5 red herrings in the Danish smørrebrød theme. Sample standard deviations show
    the spread of $A_{\mathrm{cell}}$ (set to 0 for equal values). For
    $A_{\mathrm{puzzle}}$, the mean values include all information. Sizes marked in grey
    are not evaluated. o3-mini performs better than GPT-4o mini for all evaluated sizes.
  }
    \label{fig:scores}
\end{figure*}

Fig.~\ref{fig:scores} shows the mean performance metrics of o3-mini and GPT-4o mini for
different puzzle sizes and 5 red herrings. Based on the metrics, we see that 2×3 and 4×5
are suitably difficult sizes for GPT-4o mini and o3-mini, respectively, as their mean
puzzle-level accuracies, $\overline{A_{\mathrm{puzzle}}}$, are $0.36\pm0.05$ and
$0.42\pm0.05$, respectively (with one $\sigma$ uncertainties).
$\overline{A_{\mathrm{cell}}}$ for the two models is $0.70\pm0.03$ and $0.66\pm0.04$,
respectively. An almost correct response that permutes the objects could get
$A_{\mathrm{cell}}=0$. This rarely happens in practice, as shown in
Appendix~\ref{app:best_cell}.

To get an overall comparison score, we compute the t-statistic between scores for all
puzzle sizes. We start by computing the difference in puzzle-level accuracy means,
$\Delta\overline{A_{\mathrm{puzzle}}}$, for each puzzle size evaluated by both LLMs (as
illustrated in Appendix~\ref{app:model_comparison}). Then, we take the mean of all the
differences across the puzzle sizes,
$\overline{\Delta\overline{A_{\mathrm{puzzle}}}}=0.47\pm0.04$ and a $t$-statistic of 13.
This shows that o3-mini performs significantly better than GPT-4o mini on these puzzles.
Almost half the puzzles were only solved by o3-mini. \footnotetext{E.g. we assume a
preference of \texttt{left\_of} over \texttt{just\_left\_of} for
$N_{\mathrm{objects}}=2$ across languages.}

\subsection{Red Herring Impact}
\label{sec:red_herring_impact}

To examine the effect of red herrings, we compare metrics with o3-mini for 0, 1 and 5
red herrings. For 0 vs. 1 red herring, we get
$\overline{\Delta\overline{A_{\mathrm{puzzle}}}}=0.009\pm0.003$ and $t=2.99$, and so,
adding a red herring slightly increases difficulty (see Appendix~\ref{app:1rh} for more
details).

\begin{figure}[h]
	\centering
    \resizebox{0.86\hsize}{!}{\includegraphics[trim={0 0.45cm 0 0.4cm},clip]{
      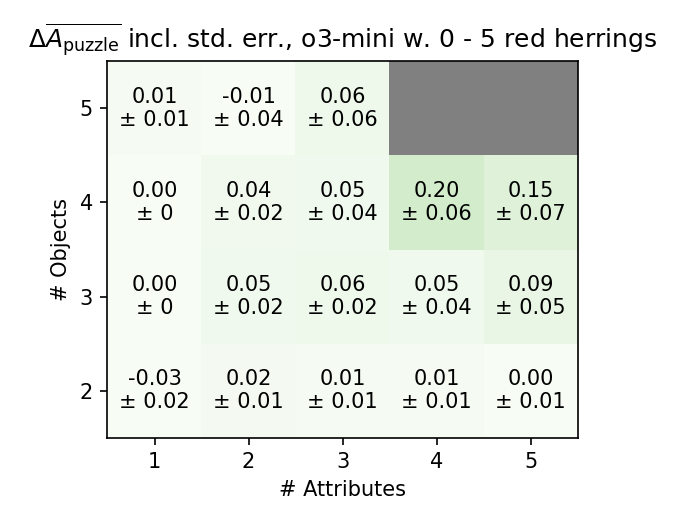}
    }
	\caption{
    $\Delta\overline{A_{\mathrm{puzzle}}}$ for o3-mini with 0 vs. 5 red herrings for 100
    puzzles in the Danish smørrebrød theme. Using 5 red herrings gives a $>2\sigma$
    decrease in $\overline{A_{\mathrm{puzzle}}}$ for sizes 3×2, 3×3, 3×5, 4×4, and 4×5.
  }
    \label{fig:rh_comparison}
\end{figure}

If we add 5 red herrings instead,
$\overline{\Delta\overline{A_{\mathrm{puzzle}}}}=0.032\pm0.007$ and $t=4.77$. Going from
0 to 5 red herrings decreases $\overline{A_{\mathrm{puzzle}}}$ by $4\pm1$ times as much
as adding 1. Fig.~\ref{fig:rh_comparison} shows that the impact appears in large
puzzles, with $\Delta\overline{A_{\mathrm{puzzle}}}=0.15\pm0.07$ for 4×5 with 5 red
herrings.

Small puzzles are easy to o3-mini with or without red herrings. Using 5 red herrings has
little impact on GPT-4o mini;
$\overline{\Delta\overline{A_{\mathrm{puzzle}}}}=0.019\pm0.005$ and
$\Delta\overline{A_{\mathrm{puzzle}}}=0.06\pm0.07$ for 2×3. Adding red herrings can be a
simple alternative to increasing puzzle size for reasoning models.

\subsection{Language Comparison}
\label{sec:language_comparison}

We compare evaluation metrics in Table~\ref{tab:theme_scores} between themes and two
languages: English, a high resource language, and Danish, a medium resource language.
$\overline{A_{\mathrm{puzzle}}}$ and $\overline{A_{\mathrm{cell}}}$ vary by $<2\sigma$
-- both for Danish vs. English house-themed puzzles and for the Danish houses vs.
smørrebrød themes. The means and sample standard deviations are close to 0.5 for both
metrics, indicating that individual puzzle metrics often vary wildly between the
possible values from 0 to 1. Logical reasoning ability appears generalisable even for a
culture-specific theme, and so, we use the houses theme for MultiZebraLogic, as it is
easier to translate.

\subsection{Clue Type Difficulty}
\label{sec:clue_type_difficulty}

To measure effect of clue and red herring types on difficulty, we compare their
frequencies to $A_{\mathrm{cell}}$. For each puzzle size, we fit to $A_{\mathrm{cell}}$
as a function of clue type frequencies using linear regression. The model coefficients
show the importance of clue types. We normalise them, so their absolute values sum to 1,
and flip the sign to arrive at the clue type difficulty. Thus, the higher the difficulty
of a clue type, the more that clue type reduces the cell accuracy when present:

\begin{align}
  \mathrm{difficulty}_{\mathrm{clue\ type}} =
    - \frac{\mathrm{coefficient}_{\mathrm{clue\ type}}}{\sum|\mathrm{coefficient}|}.
\end{align}

Section~\ref{sec:red_herring_impact} shows that red herrings contribute negatively to
accuracy, but if we keep the number of red herrings per puzzle constant, no red herring
type particularly confuses o3-mini compared to the rest. There is also no clear pattern
in clue type difficulties among the real clues across puzzle sizes when testing on 100
puzzles per size. See Appendix~\ref{app:clue_diff} for more details.

\section{Discussion and Perspectives}
\label{sec:discussion_and_perspectives}

For o3-mini with medium reasoning effort, ZebraLogicBench found an
$\overline{A_{\mathrm{puzzle}}}$ of 88~\% and an $\overline{A_{\mathrm{cell}}}$ of
90.4~\% for large puzzles of sizes 4×5, 5×3, 4×6, 5×4 and 6×3. This is higher than our
accuracies for 4×5 (42~\% and 70~\%) and 5×3 (73~\% and 80~\%) in Fig.~\ref{fig:scores}.
Our puzzles are more difficult, and Fig.~\ref{fig:rh_comparison} shows that this can be
fully explained by red herrings as they decrease $\overline{A_{\mathrm{puzzle}}}$ by
15$\pm$7~\% for 4×5 puzzles.

Several corrections and adjustments have been applied since the analysis of this paper,
which could slightly improve model performance. For example, only using the word
football instead of soccer in English. We describe the changes in
Appendix~\ref{app:adjustments}. With more advanced LLMs, evaluating broader or more
advanced reasoning skills could be useful. We suggest more puzzle and clue types in
Appendix~\ref{app:suggestions}.

\section{Conclusion}
\label{sec:conclusion}

We have published MultiZebraLogic datasets for benchmarking logical reasoning, and code
for dataset generation. New languages or themes can be added as input for easy adaption.
o3-mini can solve larger puzzles than GPT-4o mini, so for evaluation of reasoning
models, we include 4×5 puzzles, and for other models, 2×3 puzzles. We always include 5
red herrings (and publish their indices), as this causes a
$\overline{A_{\mathrm{puzzle}}}$ drop of 15$\pm$7~\% for o3-mini with 4×5 puzzles.
Logical reasoning appears generalisable for o3-mini on 4×5 puzzles across Danish and
English, and across the classic houses theme compared to the culture-specific smørrebrød
theme. The puzzle generation algorithm prefers more informative clue types, but we find
no clear correlation between included clue or red herring types and $A_{\mathrm{cell}}$.
The published dataset contains 128 puzzles for training (as few-shot examples) and 1024
for testing for sizes 2×3 and 4×5 in 9 languages.

\section{Acknowledgements}
\label{sec:acknowledgements}

We are very grateful to everyone who helped review the translations and language
configuration files\footnote{Annika Simonsen, Gardar Ingvarsson Juto, Lars Bungum,
Mathias Stenlund, Jenny Kunz and Eike Güldenring.}. We thank the EU Horizon project
TrustLLM (grant agreement number 101135671) and Danish Foundation
Models\footnote{https://www.foundationmodels.dk/} for funding this project.

\clearpage
\section{Bibliographical References}
\label{sec:reference}

\bibliographystyle{lrec2026-natbib}
\bibliography{bib}

\section{Language Resource References}
\label{sec:language_resource_references}

\bibliographystylelanguageresource{lrec2026-natbib}
\bibliographylanguageresource{languageresource}

\clearpage
\appendix
\begin{appendices}

\section{Advanced English Houses Example}
\label{app:advanced_puzzle}

A longer, more advanced zebra puzzle example compared to Figure~\ref{fig:simple_puzzle}, can be found in Figure~\ref{fig:advanced_puzzle}.

\begin{figure}
\scriptsize
\begin{verbatim}
A row of houses have numbers 1 to 4 from left to 
right.

In each house lives a person with unique attributes 
in each of the following categories:

Jobs: baker, nurse, shop assistant and teacher.
Pets: budgerigar, cat, dog and rabbit.
Drinks: coffee, juice, milk and tea.
Hobbies: board games, handball, soccer and tennis.
Favourite fruits: apple, blackcurrant, orange and 
wild strawberry.

We also know the following:

1. The person with a master's degree in mathematics 
   does not live in house no. 1.
2. The teacher lives to the immediate right of the 
   coffee drinker.
3. The shop assistant lives to the immediate right 
   of the budgie owner.
4. The rabbit owner does not live between the coffee 
   drinker and the juice drinker, and they are three 
   different people.
5. The dog owner does not like apples.
6. The person who owns a cactus often sails.
7. There are 2 houses between the nurse and the baker.
8. The tea drinker does not live next to the person 
   who loves blackcurrants, and they are different 
   people.
9. There is one house between the coffee drinker and 
   the milk drinker.
10. There are many cars on the street.
11. There are 2 houses between the milk drinker and 
    the tea drinker.
12. The nurse lives next to the dog owner.
13. There is one house between the person who plays 
    board games and the person who plays handball.
14. The person who plays football lives next to the 
    person who plays board games.
15. There are 2 houses between the person who plays 
    football and the person who loves blackcurrants.
16. The person with a tattoo does not live in house 
    no. 3.
17. The milk drinker is good friends with the person 
    with a pet that is old for its species.
18. There is one house between the cat owner and the 
    person who loves oranges.

Who has which attributes and lives in which house?

Please submit your answer as a JSON dictionary in the 
format below. Each row must begin with object_X where 
X is the house number. Each column represents a 
category, and they should be in the same order as in 
the list of categories above.

{
    "object_1": [
        "jobs_1",
        "pets_1",
        "drinks_1",
        "hobbies_1",
        "favourite fruits_1"
    ],
    "object_2": [
        "jobs_2",
        "pets_2",
        "drinks_2",
        "hobbies_2",
        "favourite fruits_2"
    ],
    "object_3": [
        "jobs_3",
        "pets_3",
        "drinks_3",
        "hobbies_3",
        "favourite fruits_3"
    ],
    "object_4": [
        "jobs_4",
        "pets_4",
        "drinks_4",
        "hobbies_4",
        "favourite fruits_4"
    ]
}
\end{verbatim}
\caption{
  A zebra puzzle with 4 objects and 5 attributes for each object (4x5). Five red
  herrings are also included in the list of clues.
}
\label{fig:advanced_puzzle}
\end{figure}

\section{Clue Type Examples}
\label{app:clue_examples}

Table~\ref{tab:clue_examples} shows an example of each clue type and
Table~\ref{tab:rh_examples} shows an example of each red herring type.

\begin{table}[h]
    \centering
    \small
    \begin{tabular}{p{0.39\linewidth} | p{0.5\linewidth}}
        Clue type & Example \\
        \hline
        \texttt{found\_at} &
          The person who plays board games lives in house no. 2. \\
        \texttt{not\_at} &
          The science fiction reader does not live in house no. 1. \\
        \texttt{same\_object} &
          The police officer reads crime novels. \\
        \texttt{not\_same\_object} &
          The dog owner does not like apples. \\
        \texttt{next\_to} &
          The zebra owner lives next to the person who loves strawberries. \\
        \texttt{not\_next\_to} &
          The person who boulders does not live next to the person who loves
          blackcurrants, and they are different people. \\
        \texttt{just\_left\_of} &
          The teacher lives to the immediate left of the rabbit owner. \\
        \texttt{just\_right\_of} &
          The teacher lives to the immediate right of the coffee drinker. \\
        \texttt{left\_of} &
          The rabbit owner lives to the left of the person who plays board games. \\
        \texttt{right\_of} &
          The Brit lives to the right of the romance reader. \\
        \texttt{between} &
          The person who loves blackcurrants lives between the police officer and the
          person who loves wild strawberries. \\
        \texttt{not\_between} &
          The rabbit owner does not live between the coffee drinker and the juice
          drinker, and they are three different people. \\
        \texttt{one\_between} &
          There is one house between the Norwegian and the police officer. \\
        \texttt{multiple\_between} &
          There are 2 houses between the nurse and the baker. \\
    \end{tabular}
    \caption{An example clue for each clue type using the English houses theme.}
    \label{tab:clue_examples}
\end{table}

\begin{table}[h]
    \centering
    \small
    \begin{tabular}{p{0.39\linewidth} | p{0.5\linewidth}}
        Red herring type & Example \\
        \hline
        \texttt{same\_herring} &
          The person who loves wild strawberries loves physics. \\
        \texttt{next\_to\_herring} &
          The Dutchman lives next to the person with a bike. \\
        \texttt{double\_herring} &
          The person who owns a cactus often sails. \\
        \texttt{fact} &
          Snails are molluscs. \\
        \texttt{object\_fact} &
          The shop assistant knows that several of the houses have a green door. \\
        \texttt{friends} &
          The person who boulders is good friends with the person who plays video
          games. \\
        \texttt{herring\_found\_at} &
          The person who has been to Canada lives in house no. 3. \\
        \texttt{herring\_not\_at} &
          The person with a master's degree in mathematics does not live in house no. 1.
    \end{tabular}
    \caption{
      An example of each red herring type in the English houses theme. Some red herrings
      may sound informative, but they are all irrelevant to the solving process.
    }
    \label{tab:rh_examples}
\end{table}

\section{Clue Type Frequency}
\label{app:clue_frequency}

Clues are randomly generated, but only included when useful, and this affects the
frequencies of clue types. The number of clues may also vary between puzzles generated
with the same inputs. To compare clue type frequencies, we count and normalise them in
each puzzle, so the frequencies sum to 1. Then, we take the mean across puzzles of the
same size (same $N_{\mathrm{objects}}$ and $N_{\mathrm{attributes}}$).

\begin{sidewaysfigure*}
    \centering
    \resizebox{\hsize}{!}{\includegraphics{
      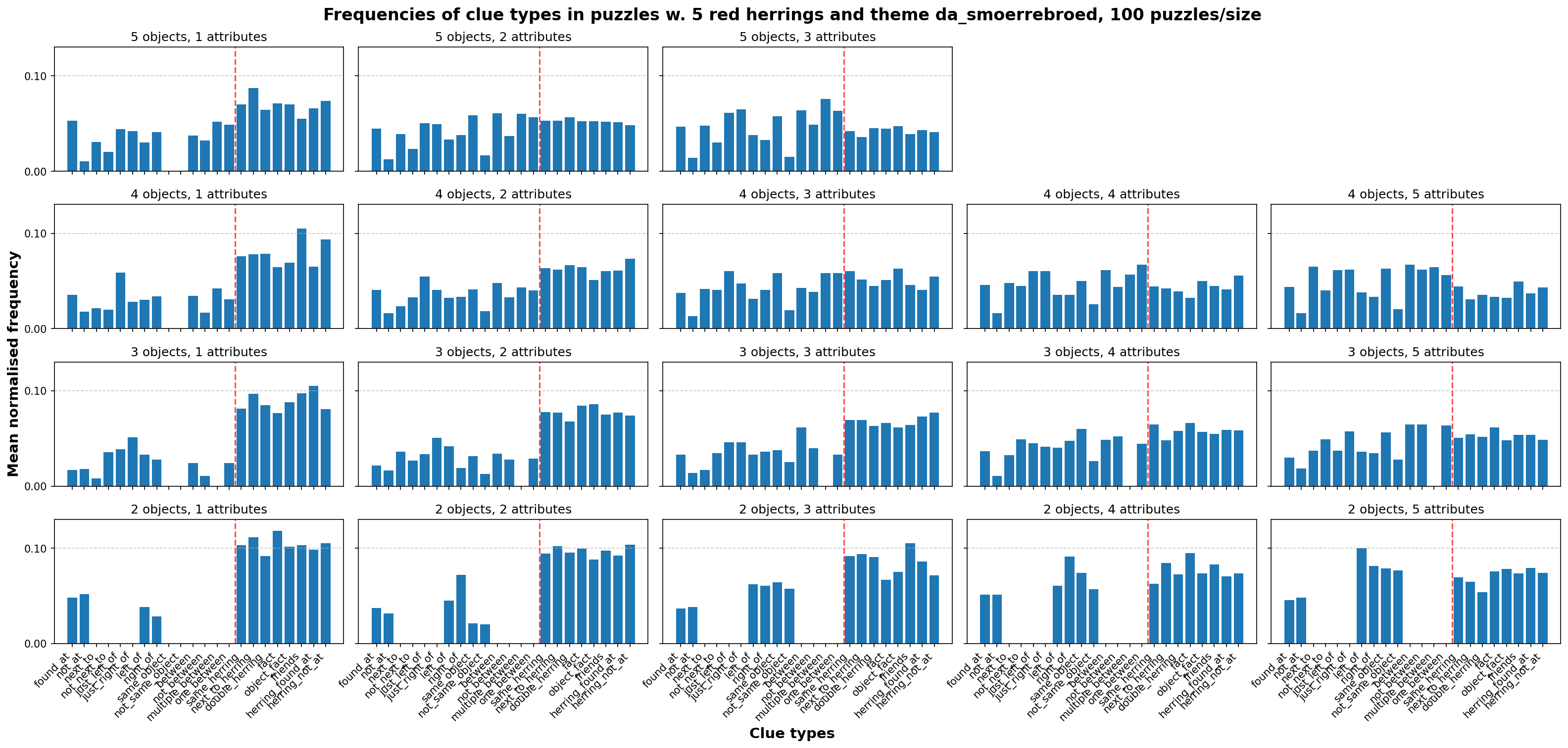}
    }
    \caption{
      Mean normalised frequencies of all clue types in puzzles with the Danish
      smørrebrød theme and 5 red herrings. To the right of the red line, all 'clues' are
      red herrings. Some clue types are only used above certain puzzle sizes -- see
      Table~\ref{tab:clue_types}. Frequently selected clues are typically more
      informative.
    }
    \label{fig:clue_freq}
\end{sidewaysfigure*}

Fig.~\ref{fig:clue_freq} shows the mean normalised frequencies for 100 puzzles with 5
red herrings. Naturally, the herrings are relatively frequent for small puzzles that
require few real clues. For real clues, the frequencies are connected to their
usefulness. For example, \texttt{not\_same\_object} is relatively rare for most puzzle
sizes, as it only excludes one link between attributes. \texttt{not\_between}-clues
connect 3 objects and fully include the \texttt{not\_same\_object}-clue -- this makes
them more informative and more common.

To change frequencies of clue types or red herring types, selection weights can be
adjusted. These are equal per default.

\section{Evaluation Details}
\label{app:evaluation}

When evaluating the models, if the API returns an \texttt{InternalServerError},
\texttt{APIError}, \texttt{APIConnectionError}, \texttt{RateLimitError},
\texttt{RateLimitError}, we wait 5 seconds and try again up to 4 more times, as these
errors do not depend on puzzle difficulty, unlike, e.g., \texttt{APITimeoutError}. For
continued errors or other error types, we treat them as a wrong solution.

\section{Uncertainty Calculation}
\label{app:std}

We will generally propagate uncertainties $\sigma$ for a function $f(a,b,...)$ using
\begin{align}
  \sigma_{f(a,b,...)} =
    \sqrt{
      \sigma_a^2\left(\frac{\partial f}{\partial a}\right)^2
      + \sigma_b^2\left(\frac{\partial f}{\partial b}\right)^2+...
    }.
\end{align}

One standard deviation corresponds to a confidence interval of 68~\% and two corresponds
to 95~\%. The sample standard deviation of the Bernoulli-distributed puzzle-level
accuracies, $A_{\mathrm{puzzle}}$, is:

\begin{align}
  \sigma_{A_{\mathrm{puzzle}}} &=
    \sqrt{\overline{A_{\mathrm{puzzle}}}*(1-\overline{A_{\mathrm{puzzle}}})}.
\end{align}

The sample standard deviation of cell-wise accuracies, $A_{\mathrm{cell}}$ is computed
as:

\begin{align}
  \sigma_{A_{\mathrm{cell}}} &=
    \sqrt{
      \frac{\sum_i|A_{\mathrm{cell},\ i}
      - \overline{A_{\mathrm{cell}}}|^2}{N_{\mathrm{puzzles}}-1}
    }.
\end{align}

To get the standard deviation of the mean scores (standard error of the mean), we divide
by $\sqrt{N_{\mathrm{puzzles}}}$:

\begin{align}
  \sigma_{\overline{A}} &=
    \frac{\sigma_{A}}{\sqrt{N_{\mathrm{puzzles}}}}.
  \label{eq:std_of_mean}
\end{align}

The standard deviation of the difference in means, $\Delta\overline{A}$, is computed as

\begin{align}
  \sigma_{\overline{A}} = \sqrt{\sigma_{A_i}^2+\sigma_{A_j}^2}
\end{align}

for models $i$ and $j$. To do this, we assume that scores can be treated as independent,
although the models can actually be evaluated on the same puzzles. The standard
deviation of the mean difference in means, $\overline{\Delta\overline{A}}$, is

\begin{align}
  \sigma_{\overline{\Delta\overline{A}}} &=
    \sqrt{
      \frac{\sum_i|(\Delta\overline{A})_i
      - \overline{\Delta\overline{A}}|^2}{N_{\mathrm{evaluated\ sizes}}-1}
    }.
\end{align}

The t-statistic (difference in units of standard deviations) is then

\begin{align}
  t = \frac{\overline{\Delta\overline{A}}}{\sigma_{\overline{\Delta\overline{A}}}}.
\end{align}

\section{Best Permuted Cell-Wise Accuracies}
\label{app:best_cell}

If a model correctly connects attributes, but switches the object numbers, this is
punished harder by $A_{\mathrm{cell}}$ than if attributes were switched within a
category. To notice if this happens, we check the best permuted cell-wise accuracy,
$A_{\mathrm{best\ cell}}$, which is the maximum cell-wise accuracy for all object
permutations. This is always equal to or higher than $A_{\mathrm{cell}}$.

The difference is not significant for responses from o3-mini on 4×5 puzzles with 5 red
herrings in the Danish smørrebrød theme. $\overline{A_{\mathrm{best\ cell}}}$ values are
generally a bit higher for GPT-4o mini with $\overline{A_{\mathrm{best\
cell}}}-\overline{A_{\mathrm{cell}}}=0.11\pm0.4$ for 2×3 puzzles. If the effect is major
for some LLMs, $A_{\mathrm{best\ cell}}$ could be considered as an extra metric for
comparison.

\section{Model comparison}
\label{app:model_comparison}

\begin{figure}[h]
	\centering
    \resizebox{0.99\hsize}{!}{\includegraphics[trim={0 0.3cm 0 0.4cm},clip]{
      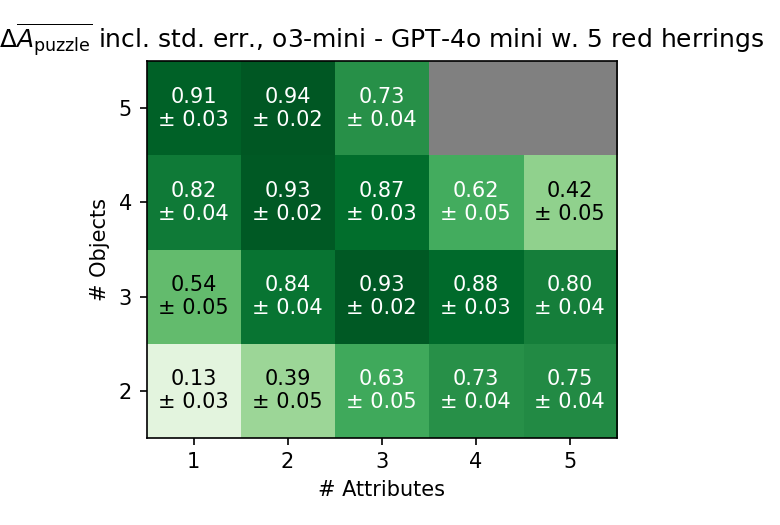}
    }
    \resizebox{0.99\hsize}{!}{\includegraphics[trim={0 0.45cm 0 0},clip]{
      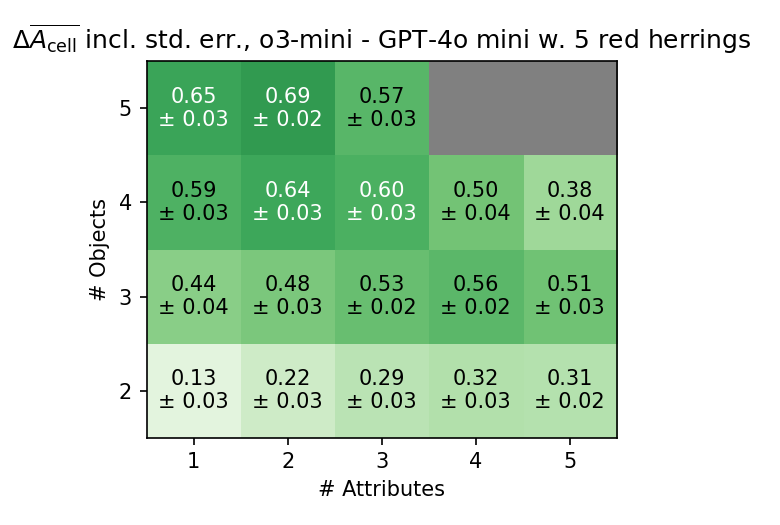}
    }
	\caption{
    Difference in mean score between o3-mini and GPT-4o mini for 100 puzzles with 5 red
    herrings in the Danish smørrebrød theme. The upper plot shows puzzle-level
    accuracies and the lower shows cell-wise accuracies. The uncertainties show the
    standard deviations of the differences in mean scores.
  }
    \label{fig:model_comparison}
\end{figure}

In Fig.~\ref{fig:model_comparison}, for each puzzle size evaluated by both models, we
take $\Delta\overline{A_{\mathrm{puzzle}}}$ and $\Delta\overline{A_{\mathrm{cell}}}$.
The figure shows that o3-mini performs better than GPT-4o mini, especially for medium
sizes such as 4×2, which are hard for GPT-4o mini but still easy for o3-mini.

\section{The Impact of One Red Herring}
\label{app:1rh}

\begin{figure}[h]
	\centering
    \resizebox{0.86\hsize}{!}{\includegraphics[trim={0 0.45cm 0 0.4cm},clip]{
      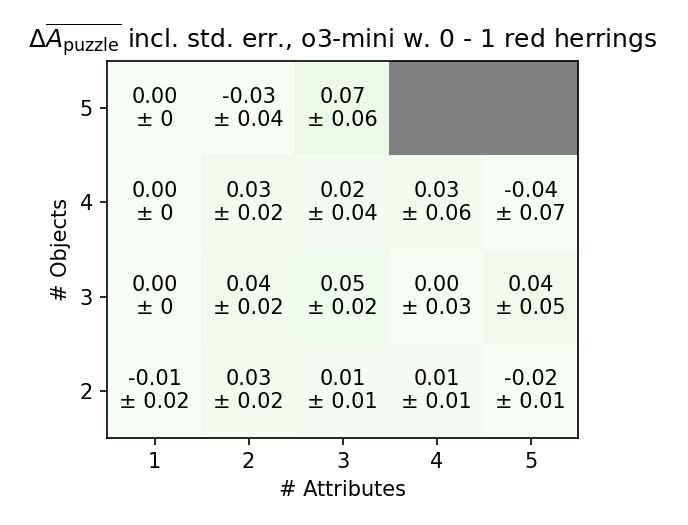}
    }
	\caption{
    $\Delta\overline{A_{\mathrm{puzzle}}}$ for o3-mini with 0 vs. 1 red herrings for 100
    puzzles in the Danish smørrebrød theme. Including 1 red herring slightly decreases
    $\overline{A_{\mathrm{puzzle}}}$, but the effect is not consistent across puzzle
    sizes.
  }
    \label{fig:1rh_comparison}
\end{figure}

Fig.~\ref{fig:1rh_comparison} shows that adding a single red herring typically decreases
$\overline{A_{\mathrm{puzzle}}}$, but the effect is very small and not significant for
most puzzle sizes -- even the largest ones, where we see the greatest effect of adding 5
red herrings in Fig.~\ref{fig:rh_comparison}.

\section{Clue type difficulties}
\label{app:clue_diff}

\begin{sidewaysfigure*}
    \centering
    \resizebox{\hsize}{!}{\includegraphics{
      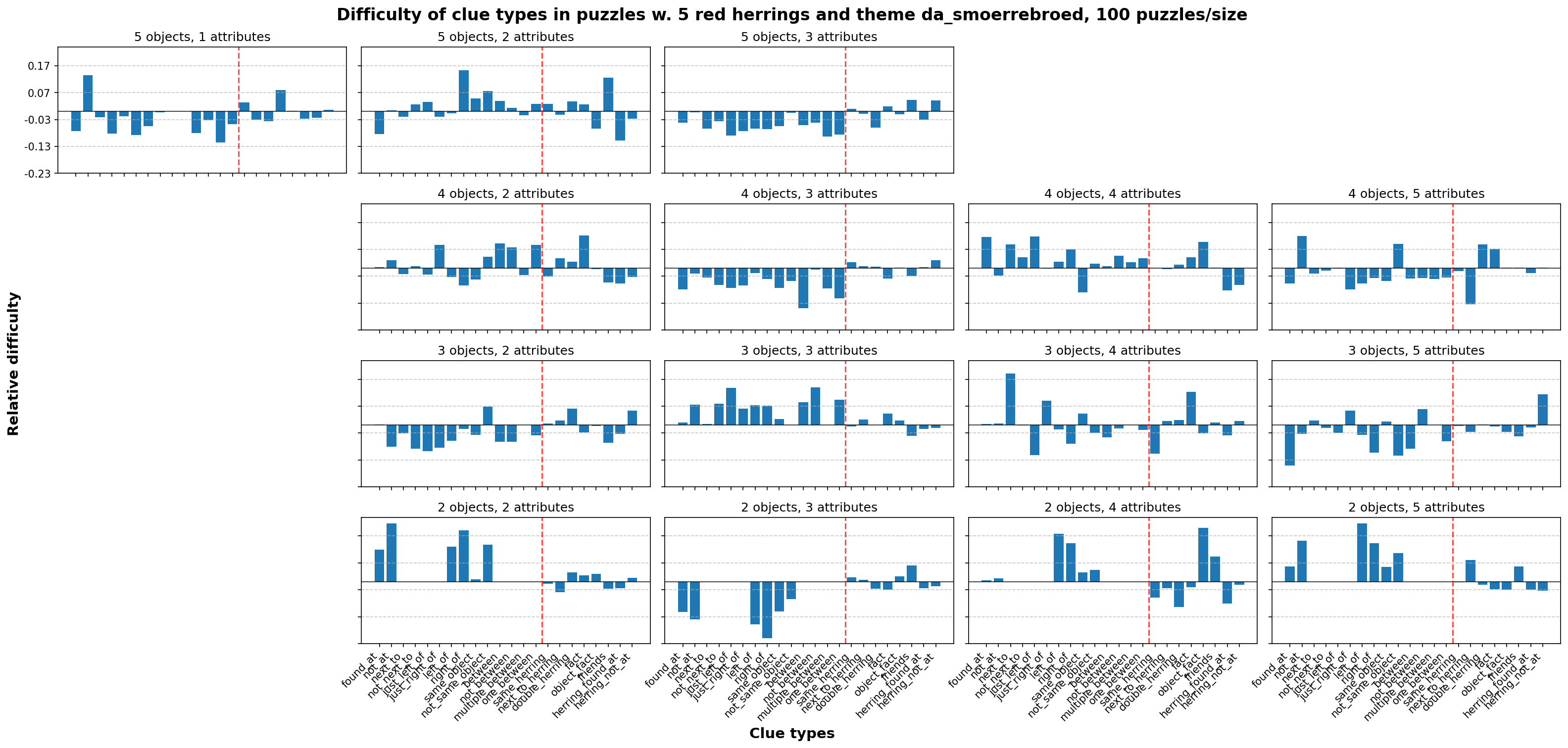}
    }
    \caption{
      Clue type difficulties as predicted contributions of clue type frequencies to
      $A_{\mathrm{cell}}$ values for o3-mini on puzzles in the Danish smørrebrød theme
      with 5 red herrings. Red herrings are on the right side of the red line. Some
      small puzzle sizes are not included, as difficulties cannot be estimated for
      constant $A_{\mathrm{cell}}$.
    }
    \label{fig:clue_diff}
\end{sidewaysfigure*}

In Fig.~\ref{fig:clue_diff}, clue type difficulties are shown for o3-mini. They show no
consistent pattern across the puzzle sizes. Clue type difficulties for o3-mini are more
accurate for large puzzles, as $A_{\mathrm{cell}}$ values are more diverse (see
Fig.~\ref{fig:scores}).

\section{Adjustments and Corrections}
\label{app:adjustments}

Multiple linguistic adjustments have been made since the results of this paper were
computed. Below we mention the most important changes.

For red herring generation, we have replaced the interest in watching football, as this
could be confused with the hobby of playing football, which is an attribute in some
puzzles. These occur together in about 11~\% of 4×5 puzzles and 3~\% of 2×3 puzzles --
both with 5 red herrings. We have replaced watching football with watching ski jumping.
We were also using the words 'soccer' and 'football' interchangeably in English, and are
now only using 'football'.

We are testing a different puzzle template including a new description of the desired
JSON format in which sorting the attributes by category is not required. If this works
well for most LLMs on Danish houses in EuroEval, it will be translated to all included
languages. Otherwise, we will consider further clarification of the rules etc.

\section{Suggested Expansions}
\label{app:suggestions}

To expand how logical reasoning is evaluated, an approach would be to use more puzzle
types. A variation of zebra puzzles could be houses on a grid instead of a linear
street. Attributes could also be non-unique or described by super-attributes (e.g. ``The
Latvian owns an animal larger than a cat'' which could be a zebra or a dog) or ordinal
attributes (e.g. ``The poetry reader owns a larger animal than the Latvian does''). Some
houses could be empty or house multiple people. One person could also have multiple
attributes in the same category.

For the current puzzle type, different clue types could be introduced, such as
``half-herrings'' that provide some useful and some useless information. For example,
``The minister's sister likes to make paintings of the baker's cat'' reveals that the
baker is the cat owner, but not which resident likes to paint, as the sister might not
live on the same street.

Other types of clues could be added for variety, such as ``The baker is either Norwegian
or has a dog'', and for all real clue types, a red herring type of a similar structure
could be created.

\end{appendices}
\end{document}